\documentclass[10pt,twocolumn,letterpaper]{article}

\makeatletter
\def\thanks#1{\protected@xdef\@thanks{\@thanks
        \protect\footnotetext{#1}}}
\makeatother

\pdfoutput=1
\usepackage{iccv}
\usepackage{times}
\usepackage{epsfig}
\usepackage{graphicx}
\usepackage{amsmath}
\usepackage{amssymb}
\usepackage{makecell}
\usepackage{multirow}
\usepackage{graphicx}
\usepackage{amsmath}
\usepackage{amssymb}
\usepackage{booktabs}
\usepackage{newfloat}
\usepackage{listings}
\usepackage{bm}
\usepackage{stfloats}
\usepackage{engord}
\usepackage{caption}
\usepackage{bbding}
\usepackage[noadjust]{cite}
\usepackage[breaklinks=true,bookmarks=false,urlcolor=black,colorlinks,citecolor=green]{hyperref}

\iccvfinalcopy % *** Uncomment this line for the final submission

\def\iccvPaperID{6718} % *** Enter the ICCV Paper ID here

\ificcvfinal\fi

\begin{document}

%%%%%%%%% TITLE
\title{\emph{Two Birds, One Stone}: A Unified Framework for Joint Learning of \\ Image and Video Style Transfers}

\author{Bohai Gu$^{1,3}$\;\;\;\;\; Heng Fan$^2$\;\;\;\;\; Libo Zhang$^{1,3\dag}$\\
$^1$ Institute of Software Chinese Academy of Sciences, Beijing, China\\
$^2$ Department of Computer Science and Engineering, University of North Texas, Denton TX, USA\\
$^3$ University of Chinese Academy of Sciences, Beijing, China\\
\thanks{$^{\dag}$Corresponding author: Libo Zhang (libo@iscas.ac.cn).}
}

\maketitle
% Remove page # from the first page of camera-ready.
\ificcvfinal\thispagestyle{empty}\fi

%%%%%%%%% ABSTRACT 

\begin{abstract}
Current arbitrary style transfer models are limited to either image or video domains. In order to achieve satisfying image and video style transfers, two different models are inevitably required with separate training processes on image and video domains, respectively. In this paper, we show that this can be precluded by introducing \textbf{UniST}, a \textbf{Uni}fied \textbf{S}tyle \textbf{T}ransfer framework for both images and videos. At the core of UniST is a domain interaction transformer ($DIT$), which first explores context information within the specific domain and then interacts contextualized domain information for joint learning. In particular, $DIT$ enables exploration of temporal information from videos for the image style transfer task and meanwhile allows rich appearance texture from images for video style transfer, thus leading to mutual benefits. Considering heavy computation of traditional multi-head self-attention, we present a simple yet effective axial multi-head self-attention (AMSA) for $DIT$, which improves computational efficiency while maintains style transfer performance. To verify the effectiveness of UniST, we conduct extensive experiments on both image and video style transfer tasks and show that UniST performs favorably against state-of-the-art approaches on both tasks. Code is available at \url{https://github.com/NevSNev/UniST}.
\end{abstract}
%-------------------------------------------------------------------------
%%%%%%%%% BODY TEXT 
\section{Introduction}
Artistic image style transfer~\cite{DBLP:conf/cvpr/GatysEB16} aims at migrating a desirable style pattern from an inference image to the origin image while preserving the original content structures. Although CNNs based methods have been well studied in this field~\cite{huang2017arbitrary,li2017universal,sheng2018avatar,li2019learning}, they fail to capture the long-range interaction between the style and content domains, which may result in suboptimal results. 

Recently, owing to the ability to model long-range dependencies, Transformers~\cite{vaswani2017attention} have shown excellent performance in a wide range of tasks including style transfer. For example, Stytr$^2$~\cite{deng2021stytr} introduces a pure transformer network to deal with image style transfer. However, pixel-level self-attention brings additional computational complexity, resulting in lower efficiency.

Unlike image style transfer, video style transfer  brings in new challenges of preserving temporal consistency between stylized video frames. To achieve style transfer on the video domain, a feasible solution is to adapt existing image-based style transfer methods (\textit{e.g.}~\cite{li2019learning,liu2021adaattn}) by re-training them with modification in architecture and/or loss functions. Despite simplicity, this domain adaption requires another repetitive and tedious training process, resulting in resource waste to some extent. Some other methods (\textit{e.g.} ~\cite{deng2021arbitrary,DBLP:journals/corr/abs-2207-04808}) directly adopt the same model from video to image, but the results are somewhat visually flawed.   

\begin{figure}[t] 
\centering 
\includegraphics[width=0.45\textwidth]{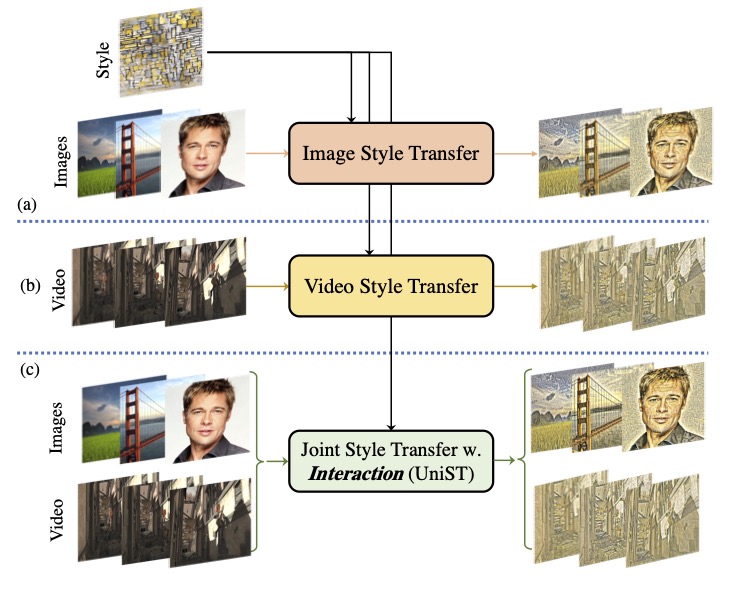}     
\caption{Comparison between single domain (image or video) style transfer and our joint style transfer.}
\label{img:compare} 
\vspace{-1em}
\end{figure}

To solve the above issues, we present a \textbf{Uni}fied \textbf{S}tyle \textbf{T}ransfer framework, termed \textbf{UniST}, for both images and videos. The proposed network leverages the local and long-range dependencies jointly. More specifically, UniST first applies CNNs to generate tokens, and then models long-range dependencies to excavate domain-specific information with domain interaction transformer ($DIT$). Afterwards, $DIT$ sequentially interacts contextualized domain information for joint learning. Considering that the vanilla self-attention suffers from a heavy computational burden, we are inspired by axial attention~\cite{wang2020axial} and develop the Axial Multi-head Self-Attention (AMSA) mechanism to calculate attention efficiently for either images or video input. To our best knowledge, our approach is the first unified solution to handle both image and video style transfers simultaneously. 

To verify the effectiveness of our approaches, we carry out extensive experiments on ImageNet~\cite{deng2009imagenet} and MPI~\cite{butler2012naturalistic} for image and video field respectively. The results demonstrate that our unified solution can achieve better performance than current state-of-the-art image-based and video-based algorithms, evidencing its superiority and efficiency.

In summary, we make the following contributions in this work: (1) We propose a new joint learning framework for arbitrary image and video style transfers, in which two tasks can benefit from each other to improve the performance. To our best knowledge, this is the first work towards a unified solution with joint interaction. (2) We develop the Axial Multi-head Self-Attention mechanism to address computational complexity and adapt to tokens from image and video input. (3) Extensive experiments on both image and video style transfer tasks demonstrate the effectiveness of our approach compared with state-of-the-art methods.

\begin{figure*}[t] 
    \centering 
\includegraphics[width=0.95\textwidth]{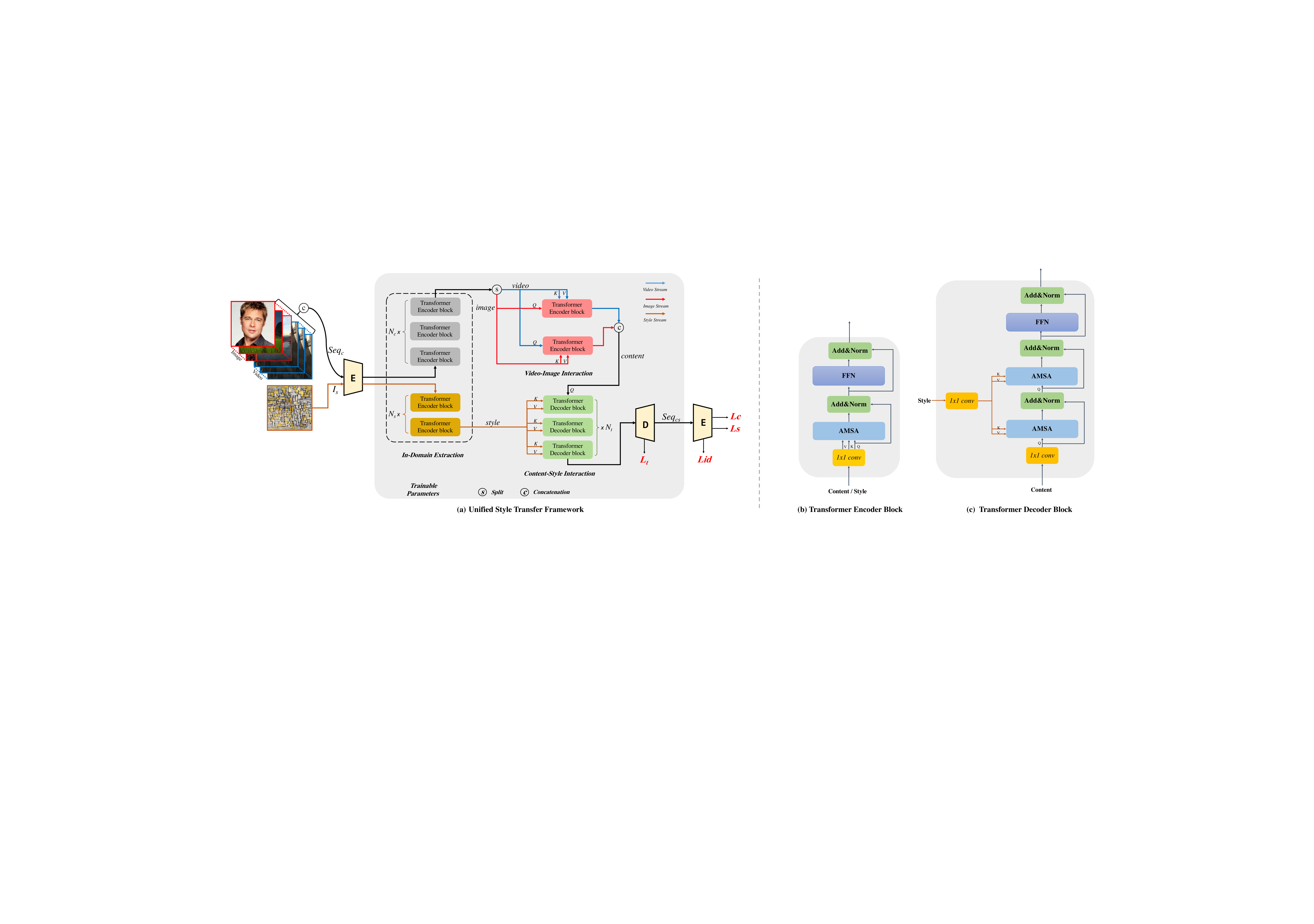}
    \caption{ (a) Overview of the UniST, where the $E$ is the VGG-19 network (pretrained and fixed) and $D$ is the CNN decoder with a symmetric structure of VGG-19.  $\mathcal{L}_c$, $\mathcal{L}_s$, $\mathcal{L}_{id}$ and $\mathcal{L}_t$ are content loss, style loss, identity loss and temporal loss; (b) The structure of improved transformer encoder block; (c) The structure of improved transformer decoder block.}
    \label{img:Overall Framework} 
\vspace{-1em}
\end{figure*}

%-------------------------------------------------------------------------
\section{Related Work}
\label{sec:related work}
\textbf{Image Style Transfer}
\label{sec:related work:Image Style Transfer}
CNNs based style transfer models are widely applied in the image field. Gatys \etal~\cite{DBLP:conf/cvpr/GatysEB16} apply the CNN model to iteratively generate stylized outputs. Johnson \etal~\cite{johnson2016perceptual} adopt an end-to-end model to accomplish real-time style transfer for the specific style. More generally, fast arbitrary style transfer is attracting enormous attention. Therefore, Huang \etal~\cite{huang2017arbitrary} achieve arbitrary style transfer by adaptively applying mean and standard deviation of style to that of content (AdaIN), which is widely adopted in image generation tasks~\cite{lin2021drafting,karras2019style} for better feature fusion. Similarly, Li \etal~\cite{li2017universal} accomplish style transfer with two transformation steps including whitening and coloring. 
Then, Sheng \etal~\cite{sheng2018avatar} design a multi-scale model combined with AdaIN and style-swap.  

Recently,~\cite{park2019arbitrary,deng2020arbitrary,liu2021adaattn,deng2021arbitrary,DBLP:conf/iccv/WuHS021,DBLP:conf/nips/ChenZWZZLXL21,luo2022consistent} introduce the self-attention mechanism to the encoder-transfer-decoder framework for better style transfer. Moreover, Deng \etal~\cite{deng2021stytr} take advantage of the Transformer's long-range dependencies while refusing to adopt the CNN's local dependencies. Meanwhile, the pure transformer is hugely computational and the position encoding needs to be presented specially. All of the above lead to slow inference speed. In contrast, UniST leverages both the Transformer's long-range dependencies and CNN's locality dependencies to build a unified framework for joint learning of image and video style transfers. After one pass of training, our framework can generate vivid image and video stylization results in real-time applications.

\textbf{Video Style Transfer}
\label{sec:related work:Video Style Transfer}
In addition to image style transfer, video style transfer presents new challenges, including both vivid stylization results and well-maintained temporal consistency. 
To this end, many previous works ~\cite{DBLP:conf/cvpr/HuangWLMJZLL17,DBLP:conf/iccv/ChenLYYH17,DBLP:conf/wacv/GaoLY020,DBLP:conf/wacv/XiaXLSCKC21} directly add optical flows consistency constraint to image style transfer solutions to enhance the inter-frame correlation. However, the optical flow requires extra complex computation, making it impractical to process high-resolution or long videos. So there emerge some works that addresses the stability issue with other approaches instead of optical flow warping. Li \etal~\cite{li2019learning} present a linear transformation module which based on content and style features. And Wu \etal~\cite{wu2020preserving} propose a SANet based framework that addresses the temporal consistency with a SSIM consistency constraint. Moreover, Deng \etal~\cite{deng2021arbitrary} learn the per-channel correlation via a transformed self-attention. On this routine, Liu \etal~\cite{liu2021adaattn} improve the temporal consistency via a modified self-attention and a cross-image similarity loss. Similarly, Wu \etal~\cite{DBLP:journals/corr/abs-2207-04808} devise a generic contrastive coherence preserving loss applied to local patches. Despite meeting multi-task domains, the results are somewhat flawed.

In this work, without using any inter-frame information like optical flows, UniST takes advantage of the unified image-video joint learning style transfer framework to facilitate video stylization effects. And the experiments demonstrate that it also achieves great temporal consistency.

\section{Methodology}
%-------------------------------------------------------------------------
\noindent
\textbf{Overall Framework.} 
Given a style image $I_s\in R^{H\times W\times3}$ and the content sequence $Seq_c \in R^{T\times H\times W\times3}$, which is the concatenation of image and video. Our framework eventually synthesizes the stylized sequence $Seq_{cs}\in R^{T\times H\times W\times3}$.
As in Figure \ref{img:Overall Framework}, our framework leverages local and long-range dependencies jointly.
Notably, the CNN encoder is not only used for local spatial information extraction, but also for tokenization. Then we use $DIT$ to accomplish two types of style transfer task jointly, which first explores context information within the content and style domains and then interacts contextualized domain information for joint learning. Below, we will detail our framework.

\subsection{Tokenization}
\label{sec:CNN Encoder CNN Decoder}
As discussed above, Deng \etal~\cite{deng2021stytr} split the input images into patches directly for transformer input with the necessary position encoding, resulting in low efficiency. In this work, we take advantage of CNNs to strengthen locality and improve the efficiency. Similar to \cite{huang2017arbitrary}, we use the pre-trained VGG-19 network~\cite{simonyan2014very} to extract feature maps of input images. Then, the $512$-dim vector at each pixel in the $relu4\_1$ layer is treated as a token for further transformer encoders. 
In this way, we combine CNNs and transformer to exploit both local and long-range dependencies. Meanwhile, there is no need to maintain the position encoding. 

\subsection{Domain Interaction Transformer}

As in Figure \ref{img:Overall Framework}, $DIT$ consists of three modules in turn, namely intra-domain extraction, video-image and style-content interaction. Specifically, the domain-specific extraction module is stacked with $N_c$, $N_s$ transformer encoder blocks for the $Seq_c$ and $I_s$ respectively. While content-style interaction  is stacked with $N_t$ transformer decoder blocks. 

\textbf{In-Domain Extraction}
\label{sec:Domain Specific Extraction}
Based on the tokenization, the local context has already been extracted. So $DIT$ further exploits domain-specific information with a number of consecutively stacked transformer encoder blocks in Figure \ref{img:Overall Framework}(b). Taking either $Seq_c$ or $I_s$ as input, the in-domain extraction module simultaneously capture the long-range information within the content and style domains. Normally, the transformer encoder block consists of a multi-head self-attention (MSA) layer and a feed-forward network (FFN). For the better efficiency, we develop the Axial Multi-head Self-Attention (AMSA) mechanism to replace MSA, combined with a $1\times1$ convolutional layer for locality strengthening. And this mechanism is applied to all of the transformer blocks mentioned below. In addition, residual connections and layer normalization are deployed after each layer. The transformer encoder block is defined as:

\begin{align}
\mathcal{S}'&= \mathcal{AMSA}(\text{Conv}(Q),\text{Conv}(K),\text{Conv}(V))+Q,\\[1mm]
\mathcal{S} &= \mathcal{FFN}(\mathcal{S}')+\mathcal{S}', 
\end{align}
where $\mathcal{S}$ is the output sequence.

\textbf{Video-Image Interaction.}
After domain-specific excavation, $DIT$ presents a symmetric module based on two transformer encoder blocks to interact contextual information between two types of content modalities. Notably, we split the input content sequence into two parts: one half is video $Seq_v$ and the other half is image $Seq_I$. As illustrated in Figure \ref{img:Overall Framework}, one of the blocks takes the $Seq_v$ as $Q$, the $Seq_I$ as $K$, $V$, and the other one does the opposite. In this way, joint learning is performed for style transfer. And two types of content sequence are concatenated after interaction.

\textbf{Content-Style Interaction.}
To finally capture the relevance between the content and the style domain, this module consists of a set of stacked transformer decoder blocks. As in Figure \ref{img:Overall Framework}(c), each transformer decoder block takes the content as $Q$ while the style as the $K$ and $V$, alternately containing two ASMA layers and one FFN layer. Similarly, layer normalization and residual connections are applied after each layer. The transformer decoder block is defined as:

\begin{small}
\begin{align}
\mathcal{S}^{''} &= \mathcal{AMSA}(\text{Conv}(Q),\text{Conv}(K),\text{Conv}(V)) + Q,\\[1mm]
\mathcal{S}' &= \mathcal{AMSA}(\text{Conv}(\mathcal{S}^{''}),\text{Conv}(K),\text{Conv}(V))+\mathcal{S}^{''},\\[2mm]
\mathcal{S} &= \mathcal{FFN}(\mathcal{S}')+\mathcal{S}'.
\end{align}
\end{small}

\begin{figure}[t] 
    \centering 
\includegraphics[width=0.45\textwidth]{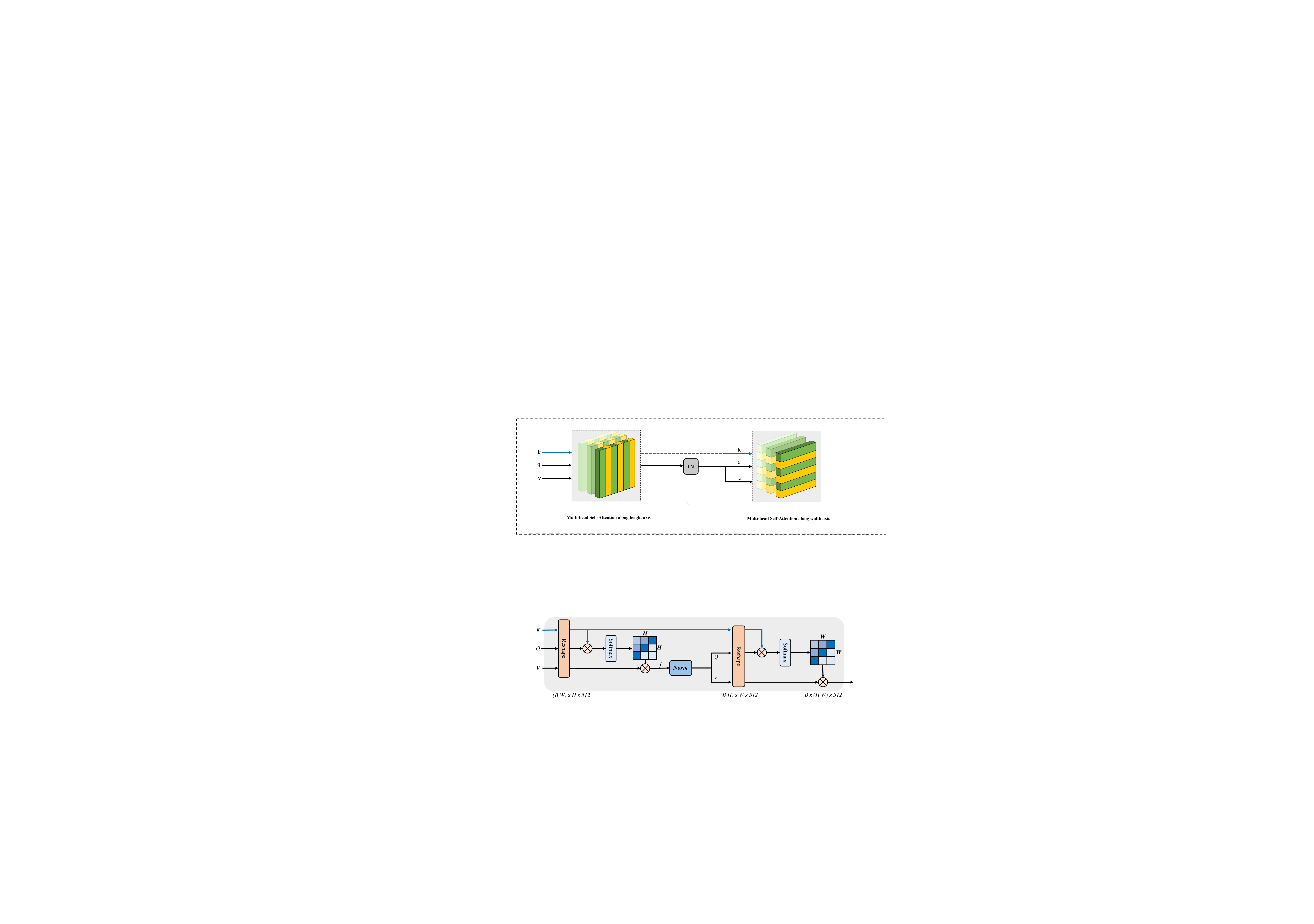}     
    \caption{The structure of the AMSA. $Norm$ here denotes the layer normalization.}
    \label{fig:axial} 
\vspace{-1em}
\end{figure}

\subsection{Axial Multi-head Self-Attention.}
Inspired by~\cite{wang2020axial}, $DIT$ computes self-attention along a separate axis, rather than in feature maps like other transformer models~\cite{deng2021stytr,DBLP:journals/corr/abs-2103-11816,DBLP:journals/corr/abs-2104-05704,DBLP:conf/iclr/DosovitskiyB0WZ21,DBLP:conf/iccv/WuXCLDY021}. The improved AMSA layer is specialized for style transfer without position encoding.

An AMSA layer consists of two MSA~\cite{vaswani2017attention} layers in total, operating sequentially along the height and width axes. As in Figure \ref{fig:axial}, we first reshape $Q$, $K$, $V$ to separate the width dimension information, then use the first MSA layer to calculate self-attention along the height axis, and the output $f$ is normalized using the layer normalization ($LN$) layer. Following almost the same process as before, except that what we separate in the second MSA layer is the height dimension. Notably, we use the previous output $f$ as $Q$ and $V$, while keeping $K$ unchanged. Experiments in Table \ref{tab:compare_fww}  verify this dedicated AMSA layer reduces the memory consumption and meanwhile improves inference efficiency. 

The reasons behind our AMSA are two-fold. First, the information contained in the previous $K$, $V$ is crucial for the second MSA layer to learn contextual dependencies either within or across domains. Besides, the $V$ of the second MSA contains the last axial information, which is critical to ensure the stability of stylization. Experiments in Figure \ref{fig:attention} show that our design effectively prevents artifacts caused by side effects of axial attention in style transfer.

\subsection{Unimodal input for inference.}
In inference, UniST takes bimodal content input and outputs both style transfer results simultaneously, achieving \emph{two birds with one stone}. Since the model has learned complementary content knowledge, UniST could also take unimodal content input into account by replacing the cross-attention with self-attention in the video-image interaction without compromising the quality of results. As in Table \ref{tab:modality input}, unimodality obtains the consistent scores with bimodality. 

\subsection{Loss functions.}
\label{sub:loss}
The overall loss function is the weighted summation of the content loss $\mathcal{L}_c$, style loss $\mathcal{L}_s$, identity loss $\mathcal{L}_\text{id}$ and temporal loss $ \mathcal{L}_t$:
\begin{equation}
     \mathcal{L} = \lambda_c \mathcal{L}_c +\lambda_s\mathcal{L}_s + \mathcal{L}_\text{id} +  \lambda_t\mathcal{L}_t,
\end{equation}
where $\lambda_c$, $\lambda_s$ and $\lambda_t$ are balancing factors. UniST uses the pre-trained VGG-19 to extract feature maps as $\phi=\{relu{1\_1},relu{2\_1},relu{3\_1},relu{4\_1}\}$. In our experiment, we use the above four layers with equal weights to calculate the loss below.

\begin{figure}[t] 
    \centering 
\includegraphics[width=0.45\textwidth]{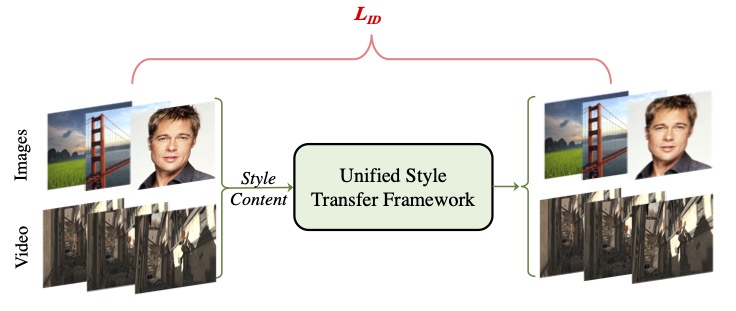}     
    \caption{Illustration of the identity loss.}
    \label{img:ID loss} 
\vspace{-1em}
\end{figure}

We use Euclidean distance~\cite{DBLP:journals/spm/DokmanicPRV15} to compute content loss $\mathcal{L}_c$:
\begin{equation}
\footnotesize
\label{equ:content}
     \mathcal{L}_c= \sum_{i=1}^{4}\left \| \phi_i(Seq_\text{cs}) -  \phi_i(Seq_{c}) \right \| _2.
\end{equation}
Following~\cite{huang2017arbitrary}, the style loss $\mathcal{L}_s$ is defined as:
\begin{equation}
\footnotesize
\label{equ:style}
\begin{aligned}
     \mathcal{L}_s &= \sum_{i=1}^{4}\big(\left \| \mu( \phi_i(Seq_\text{cs}))-   \mu(\phi_i(I_{s}) )\right \| _2 \\
     &+\left \| \sigma (\phi_i(Seq_\text{cs}))-  \sigma (\phi_i(I_{s})) \right \| _2\big),
\end{aligned}
\end{equation}
where $\mu(\cdot)$ and $\sigma(\cdot)$ denote the mean and variance of features separately. We further use the identity loss $\mathcal{L}_{id}$~\cite{park2019arbitrary} to promote more accurate content and style representation:
\begin{equation}
\footnotesize
\begin{aligned}
     &\mathcal{L}_\text{id} = \lambda_\text{id1}(\left \| Seq_\text{cc}-   Seq_{c}\right \| _2 + \left \| I_\text{ss}-  I_{s} \right \| _2)\\
     &+\lambda_\text{id2}(\sum_{i=1}^{4}\left \| \phi_i(Seq_\text{cc})-   \phi_i(Seq_{c})\right \| _2 +\left \| \phi_i(I_\text{ss})-  \phi_i(I_{s}) \right \| _2),
\end{aligned}
\end{equation}
where $\lambda_\text{id1}$ and $\lambda_\text{id2}$ both are balancing factors. $Seq_\text{cc}$/$I_\text{ss}$ denote the results synthesized from two identical content sequences or style images. Figure \ref{img:ID loss} illustrates the identity loss for better understanding.

Following the recent work AdaAttN~\cite{liu2021adaattn}, we preserve the temporal consistency via a cross-image temporal loss. This loss promotes the cosine distance $D_{cs}$ between adjacent stylization frames to be closer to the cosine distance $D_c$ between adjacent origin frames.
\begin{equation}
\footnotesize
\begin{aligned}
     &\mathcal{L}_{t} = \sum_{i=3}^{4}\phi_i(\frac{1}{N_{c_1}N_{c_2}} \sum_{m,n}\left |\frac{D_{c_1,c_2}^{m,n}}{\sum_mD_{c_1,c_2}^{m,n}}  - \frac{D_{cs_1,cs_2}^{m,n}}{\sum_mD_{cs_1,cs_2}^{m,n}}  \right |),
\end{aligned}
\end{equation}
where $D_{u,v}^{m,n} = 1 - \frac{F_u^{m}\cdot F_v^{n}}{\left \| F_u^{m} \right \| \times\left \| F_v^{n} \right \|  }$. 
$N$ is the spatial dimension of the current feature map. $D_{u,v}^{m,n,x}$ measures cosine distance, and $F^{k}$ represents the feature vector of the $k$-th entry. Note, we only adopt layer $relu3\_1$ and $relu4\_1$ to calculate $\mathcal{L}_{t}$. Meanwhile, in each training iteration, we compute the temporal loss between the consecutive video frames. Noted that temporal information is implicitly encoded in this way. 

%------------------------------------------------------------------------
\begin{figure*}[t]
\centering
\includegraphics[width=0.98\textwidth]{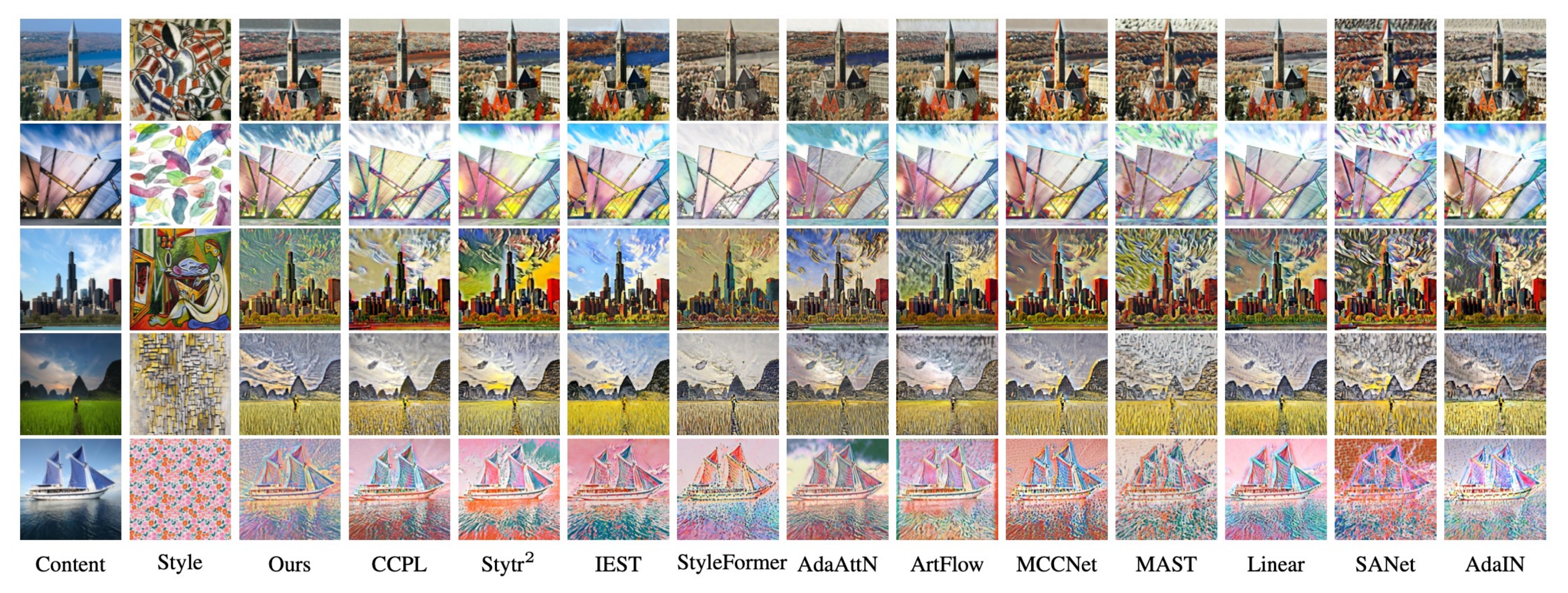}
\caption{Qualitative comparison in image style transfer. Please zoom in for better view. Additional vivid stylization results are provided in our supplementary materials.
}
\vspace{-2mm}
\label{fig:Image Qualitative Comparison}
\end{figure*}

%------------------------------------------------------------------------
\section{ Experiment}
\subsection{Implementing Details}
UniST is trained with MS-COCO~\cite{lin2014microsoft} (image), MPI~\cite{butler2012naturalistic} (video) as the content datasets and WikiArt~\cite{phillips2011wiki} as the style datasets. For $Seq_c$, the ratio of the two types of content is $1:1$, and the total number is $6$. In the training phase, all images are loaded with the resolution of $256\times256$. While in the inference phase, UniST can be applied to any resolutions. Therefore, we manage to extend the UniST to multi-granularity style transfer (see supplementary material). In experiments, we adopt the resolutions of $1024\times1024$ and $1024\times2048$ for image and video respectively. For the loss function, the balancing factors $\lambda_c$, $\lambda_s$, $\lambda_t$, $\lambda_\text{id1}$, $\lambda_\text{id2}$ are set as $0.1$, $1.5$, $90$, $0.1$, $0.5$, empirically. The number of transformer blocks $N_c,N_s,N_t$ are empirically set as $2,1,3$. Our network is trained for $50K$ iterations on a NVIDIA Tesla V100 GPU with a batch size of $3$. We use the Ranger optimizer~\cite{Ranger} with initial learning rate of $0.00005$.

%~~~~~~~~~~~~~~~~~~~~~~~~~~~~~~~~~~~~~~~
\subsection{Image Style Transfer}
We compare our method with $13$ state-of-the-art methods, including AdaIN~\cite{huang2017arbitrary}, SANet~\cite{park2019arbitrary}, Linear~\cite{li2019learning}, MCCNet~\cite{deng2021arbitrary}, MAST~\cite{deng2020arbitrary}, Artflow~\cite{an2021artflow}, AdaAttN~\cite{liu2021adaattn}, StyleFormer~\cite{DBLP:conf/iccv/WuHS021}, IEST~\cite{DBLP:conf/nips/ChenZWZZLXL21}, Stytr$^2$~\cite{deng2021stytr}, CCPL~\cite{DBLP:journals/corr/abs-2207-04808}, epsAM~\cite{cheng2023user}, MicroAST~\cite{DBLP:journals/corr/abs-2211-15313}, MCCNetV2~\cite{kong2023exploring}.

\textbf{Quantitative comparison.}
We randomly collect $17,124$ content images from ImageNet~\cite{deng2009imagenet} and $17,124$ style images from WikiArt~\cite{phillips2011wiki} which are separated from the training set to generate $17,124$ stylization results. Similar to Stytr$^2$~\cite{deng2021stytr}, we use the mean euclidean distance and the mean instance statistics difference mentioned in section \ref{sub:loss} as metrics for content preservation and stylization degree. Furthermore, we conduct the color distribution experiments by adopting color loss in DSLR~\cite{DBLP:conf/iccv/IgnatovKTVG17} and adopt the Gram matrices~\cite{DBLP:conf/cvpr/GatysEB16} for texture difference. As in Table \ref{tab:Image Quantitative Comparison}, compared with existing methods, UniST achieves the best performance in both content and style differences and obtains promising results on texture and color differences of style aspects.

\textbf{Qualitative comparison.}
\label{sec:image QC}
As in Figure \ref{fig:Image Qualitative Comparison}, AdaIN~\cite{huang2017arbitrary} transfers the style patterns but losses important content details (1st, 2nd rows). SANet~\cite{park2019arbitrary} fails to align the distribution of the style patterns, leading to the distorted object boundaries (1st, 3rd, 5th rows) and inconsistent content background structures (2nd, 4th rows). The stylization of Linear~\cite{li2019learning} is not satisfactory enough, resulting relatively light migration effects such as pink in the 5th row. As with the linear transformation, MCCNet~\cite{deng2021arbitrary} learns the correlation of each channel through the transformed self-attention, slightly improving the transfer effect, but there are serious overflow problems around the object boundaries (1st, 3rd, 4th rows). MAST~\cite{deng2020arbitrary} distorts the content background structure with excessive style transfer (3rd, 4th rows). Based on WCT~\cite{li2017universal} in our setting, Artflow~\cite{an2021artflow} leads to conspicuous vertical artifacts at the edges of the 
generated results(3rd, 5th rows). AdaAttN~\cite{liu2021adaattn} does the decent foreground style transfer, but requires further rendering of the background (3rd, 5th rows). StyleFormer~\cite{DBLP:conf/iccv/WuHS021} provides vivid style patterns, but the colors of the results are inconsistent with the style reference images (1st, 3rd, 4th rows). On the contrary, IEST~\cite{DBLP:conf/nips/ChenZWZZLXL21} presents stylization results with consistent colors, lacking more desirable style patterns (3rd, 4th, 5th rows). Stytr$^2$~\cite{deng2021stytr} fails to preserve the background structure of the content inference image (2nd, 5th rows). CCPL~\cite{DBLP:journals/corr/abs-2207-04808} transfers the style patterns with a lot of vertical artifacts (1st, 3rd, 4th rows). In contrast, based on the inductive bias of CNNs, our method captures the relevance between content and style sequences jointly and efficiently. As a result, we provide vivid stylized results with desirable style pattern details, while keeping the content structure well-maintained.

\begin{table*}[t]
  \centering
  \resizebox{\textwidth}{!}{
  \begin{tabular}{l|c c c c c c c c c c c c c c c }
    \Xhline{1pt}
    Methods & Ours &epsAM &MicroAST &MCCNetv2 &CCPL &StyleFormer &IEST &Stytr$^2$ &AdaAttN & MCCNet & Artflow & MAST & SANet  & Linear & AdaIN\\
    \Xhline{0.5pt}
     $ \mathcal{D}_C $($\downarrow$) & \bf{12.36} &20.30 &13.18 &16.15 &14.66 &16.82 &15.38 &13.67 &14.54 & 15.51 &\underline{12.55} &17.83 &17.96 & 15.04 & 18.42 \\
     $ \mathcal{D}_S $($\downarrow$) & \bf{0.46}  &1.03 &0.92 &1.06 &0.71  &0.80  &1.38  &0.50  &1.07  & 0.74  &0.85  &0.60  &\underline{0.47}  & 0.61 & 0.52\\
      $ \mathcal{TD} $($\downarrow$) & \underline{69.69} &155.54 &177.20 &143.68 &163.00 &115.84 &241.69 &89.93 &173.87 &107.91 &169.39 &94.58 &74.39 & 87.72 & \bf{66.95} \\
     $ \mathcal{CD} $($\downarrow$) &\underline{17894}  &\bf{16546} &21128 &20292 &21971  &18344  &23953  &21543  &21739  &20566  &19618  &20241  &19811  &20948 &21120 \\
    \Xhline{1pt}
  \end{tabular}
  }
  \caption{Quantitative comparison in image style transfer. The best two results are highlighted in bold and underline.}
  \vspace{-1em}
  \label{tab:Image Quantitative Comparison}
\end{table*}

\subsection{Video Style Transfer}
For video style transfer, we compare our method with four state-of-the-art methods including Linear~\cite{li2019learning}, MCCNet~\cite{deng2021arbitrary}, AdaAttN~\cite{liu2021adaattn}, and CCPL~\cite{DBLP:journals/corr/abs-2207-04808}. Note, optical flow is not used for stabilization when conducting comparison. 

\textbf{Quantitative comparison.}
Following Liu \etal~\cite{liu2021adaattn}, we adopt the official optical flow~\cite{butler2012naturalistic} to wrap the output stylized frame and compute the per-pixel difference between warped and stylized frames. Meanwhile, we use the LPIPS~\cite{DBLP:conf/cvpr/ZhangIESW18} to measure the diversity of adjacent stylized frames, with smaller values indicating better consistency.

In practise, the style input is fixed at $512\times512$, and the content input is fixed at $256\times256$.
Table \ref{tab:Optical flow error} presents optical flow error and LPIPS metrics for $20$ styles over $23$ videos of compared methods. UniST achieves the best scores in both metrics and thus has the best temporal consistency.

\begin{table}[!htb]
\renewcommand\arraystretch{1.4}
\small
\resizebox{0.48\textwidth}{!}{
\begin{tabular}{l|cccc|cccc|c}
    \Xhline{1pt}
    \multirow{2}{*}{Methods}    &\multicolumn{4}{c|}{Optical flow error($\downarrow$)}  &\multicolumn{4}{c|}{LPIPS($\downarrow$)}  &\multirow{2}{*}{$\mathcal{D}_S $($\downarrow$)}\\  
    \Xcline {2-9}{0.5pt}
                                 & Style1 & Style2 & Style3 & Mean  & Style1 & Style2 & Style3 & Mean  \\
    \Xhline{0.5pt}
       Ours                      &\bf{3.64} &\bf{6.16} & \bf{5.78} & \bf{3.86}  & \bf{1.73}  &\bf{2.05}  &\bf{2.04}  &\bf{1.79}  &\bf{13.70}  \\
    \Xhline{0.5pt}
       AdaAttN (\emph{ICCV 2021}) &4.26  & 7.09   & 6.71     &3.91              &2.26  &2.49  &2.46  &2.05 &14.56\\
    \Xhline{0.5pt}
       MCCNet (\emph{AAAI 2021})  &4.60  & 6.83   & 6.50     & 4.57             &2.13  &2.36  &2.34  &2.07 &15.21\\
   \Xhline{0.5pt} 
       Linear (\emph{CVPR 2019})  &4.23  & 6.81   & 7.10     & 4.25             &2.08  &2.27  &2.26  &2.02 &14.70\\
    \Xhline{0.5pt}
       CCPL (\emph{ECCV 2022})    &5.14  & 7.65   & 7.57     &4.90              &2.10  &2.33  &2.30   &2.06 &14.69\\
    \Xhline{1pt}
  \end{tabular}
  }
  \caption{The optical flow error ($\times10^{-2}$) and LPIPS of SOTAs using $20$ styles. Smaller values mean better temporal consistency. For clarity, only three styles are presented. Please refer to supplementary material for more details.}
  \label{tab:Optical flow error}
\end{table}

\begin{figure*}[t]
\setlength{\abovecaptionskip}{2mm}
\centering
\includegraphics[width=0.95\textwidth]{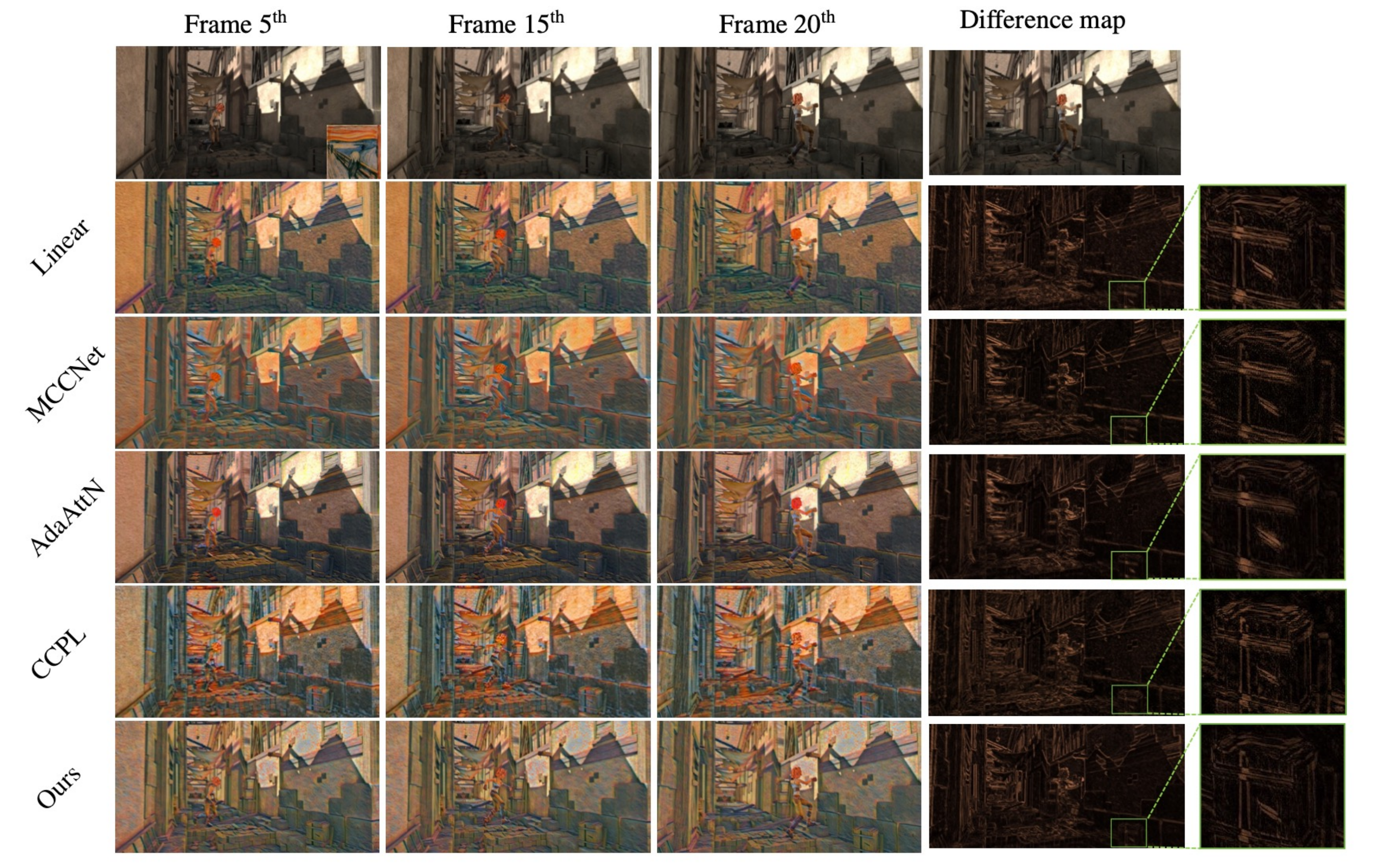}
\caption{Qualitative comparison in video style transfer.
Column $4$ shows the difference map between the adjacent frames at frame $20$. }
\label{fig:Video Qualitative Comparison}
\end{figure*}

\textbf{Qualitative comparison.}
\label{sec:video QC}
Figure \ref{fig:Video Qualitative Comparison} shows the results of the video qualitative comparison. To better verify long-term temporal consistency, we take the $5$-th, $15$-th and $25$-th frames of the example video as the reference content images, where the character in the video has large movements. The results show that all four methods visually satisfy the long-term temporal consistency. 
To be specific, Linear~\cite{li2019learning} uses the shallower feature map in video style transfer task, sacrificing the stylization effects for temporal consistency in some way.
MCCNet~\cite{deng2021arbitrary} produces distorted results with severe artifacts along object contours, where the same drawback appears in image style transfer. 
AdaAttN~\cite{liu2021adaattn} provides a well-transferred foreground leaving the background requires further rendering.
Using the same model as image style transfer, CCPL~\cite{DBLP:journals/corr/abs-2207-04808} also shows a huge number of vertical artifacts in video stylization results.
In contrast, benefiting from the joint learning framework, our video results are enhanced by the rich texture from images beyond the reference video. 
Besides, the difference map in column $4$ shows that our joint learning framework is stable enough when dealing with the motion blur. Based on the above analysis, our method can generate video results with more pleasing stylistic patterns while maintaining temporal consistency well.
Notably, when applied to image style transfer, some of these compared methods~\cite{li2019learning,liu2021adaattn} need to be retrained with extra changes and consumption, while the rest~\cite{deng2021arbitrary,DBLP:journals/corr/abs-2207-04808} use the same model directly with the flawed results. In contrast, our joint learning framework can perform well on both image and video style transfer in one go. Additional vivid stylization results are provided in our supplementary materials.

\begin{figure}[htb] 
\centering 
\includegraphics[width=0.48\textwidth]{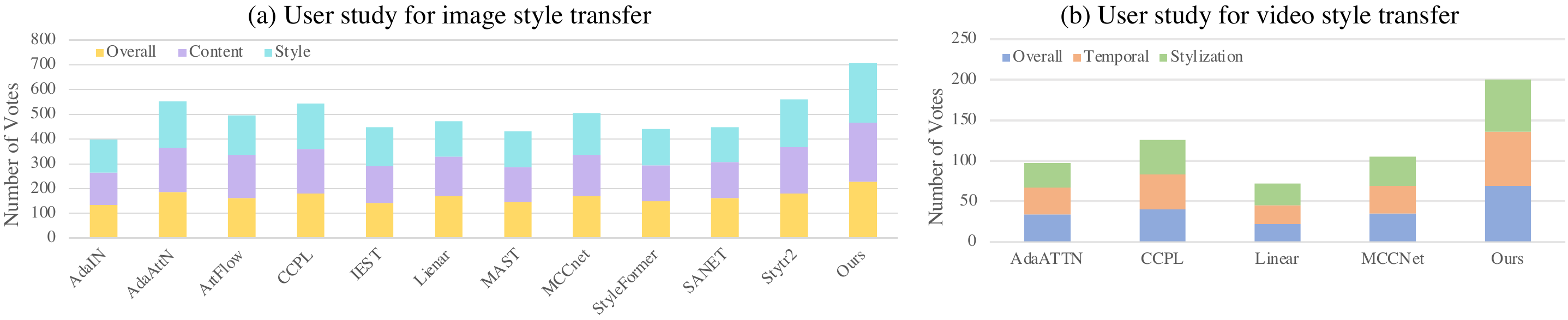} 
\caption{The user study for images (a) and video (b) style transfer. Best viewed by zooming in.}
\label{fig:user_study}
\end{figure}

%~~~~~~~~~~~~~~~~~~~~~~~~~~~~~~~~~~~~~~~~~~~~

%~~~~~~~~~~~~~~~~~~~~~~~~~~~~~~~~~~~~~~~~~~~~

\subsection{User Study}
\label{sec:User Study}
Furthermore, we carry out the user study experiment for comparison. Specifically, we use $27$ content images and $21$ style images to synthesize $567$ images in total. Given $20$ randomly combinations of content and style, the generated results obtained by $12$ image-based style transfer methods. Then, we ask $100$ participants to select their favorite one from three aspects: content preservation, style transfer, and overall preference. We collect $2,000$ votes and show results in Figure \ref{fig:user_study}(a). The results verifies the superiority of our method over other models for image style transfer. Similarly, we take $12$ videos of $50$ frames and $21$ style images to synthesize $252$ video stylized clips in total. Given $4$ random combinations of video and style, the stylization clips obtained by $5$ video-based style transfer solutions. Then, we ask another $50$ subjects to select their favorite one from three views: temporal consistency, stylization effect (considering both content preservation and stylization degree), and overall preference. We collect $200$ votes and our method is selected as the best as shown in Figure \ref{fig:user_study}(b).

For participants, there are 83 males and 67 and females (55/45 males/females for image, the other 28/22 males/females for video), aged from 23 to 42.

\subsection{Efficiency Analysis}
\label{sec:Efficiency Analysis}
In Table \ref{tab:Inference Time}, we report the inference time of our joint learning method and other approaches. Note that all the methods are run using a single TITAN XP GPU card. Although our joint learning framework consists of many stacked self-attention layers, our model can still achieves $35$ FPS at $256$px, which is comparable with attention-based methods such as AdaAttN~\cite{liu2021adaattn} and Stytr$^2$~\cite{deng2021stytr}. The main limitation of our work is the inference speed is limited when applied for high resolution input, which may hinder its usage.

\begin{table}[!htb]
\renewcommand\arraystretch{1.2}
\small
\resizebox{0.47\textwidth}{!}
{
  \begin{tabular}{l|ccccccc}
     \Xhline{1pt}
    Methods & Ours & SANet & Linear & MCCNet & Artflow & Stytr$^2$ &CCPL\\
    \Xhline{0.5pt}
      $256\times256$  & 0.028 & 0.016 & 0.011 & 0.067 & 0.144 & 0.109 &0.019\\
    \Xhline{0.5pt}
      $512\times512$  & 0.111 & 0.053 & 0.062 & 0.141 & 0.363 & 0.751 &0.061\\
     \Xhline{1pt}
    Methods  &Ours $w/$ MSA & AdaATTN & Avatar-Net & AdaIN & MAST &StyleFormer &IEST\\
    \Xhline{0.5pt}
      $256\times256$ &0.030 & 0.040 & 0.116 & 0.012  & 0.022 & 0.018 & 0.019\\
    \Xhline{0.5pt}
      $512\times512$ &0.182 & 0.127 & 0.339 & 0.037  & 0.092 & 0.062 & 0.059\\
    \Xhline{1pt}
  \end{tabular}
  }
\caption{Inference time (sec./image) comparison.}
\label{tab:Inference Time}
\vspace{-1em}
\end{table}

Notably, our AMSA mechanism is crucial for the joint learning framework to address the expensive memory consumption and huge computation complexity. In particular, the computation complexity can be reduced from $\bm{O}$($\bm{H^2}\times\bm{W^2}$) to $\bm{O}$($\bm{H^2}+\bm{W^2}$) theoretically. As shown in Table \ref{tab:compare_fww}, AMSA is much lighter compared to MSA under the same experiment setting, demonstrating the great potential for high-resolution image and long video style transfer. Although the framework consists of many self-attention layers, the UniST is still fully capable of practical applications.

\renewcommand\arraystretch{1}
\begin{table}[!ht]\scriptsize
  \centering
    \begin{tabular}{c|cc}
    \Xhline{1pt}
          & \emph{Memory(MiB)} & \emph{FLOPS (G)} \\
    \hline
    \emph{MSA}     &  1.8x10$^{4}$    &    4.29      \\
    \emph{AMSA}      &   1.1x10$^{4}$     &   0.27   \\
    \Xhline{1pt}
    \end{tabular}
    \caption{ Efficiency of AMSA by comparison with MSA.}
  \label{tab:compare_fww}
\end{table}

\subsection{Ablation Study}
\label{sec:Ablation Study}

%~~~~~~~~~~~~~~~~~~~~~~~~~~~~~~~~~~~~~~~~~~~~

\begin{figure}[hb]
\centering
\includegraphics[width=0.45\textwidth]{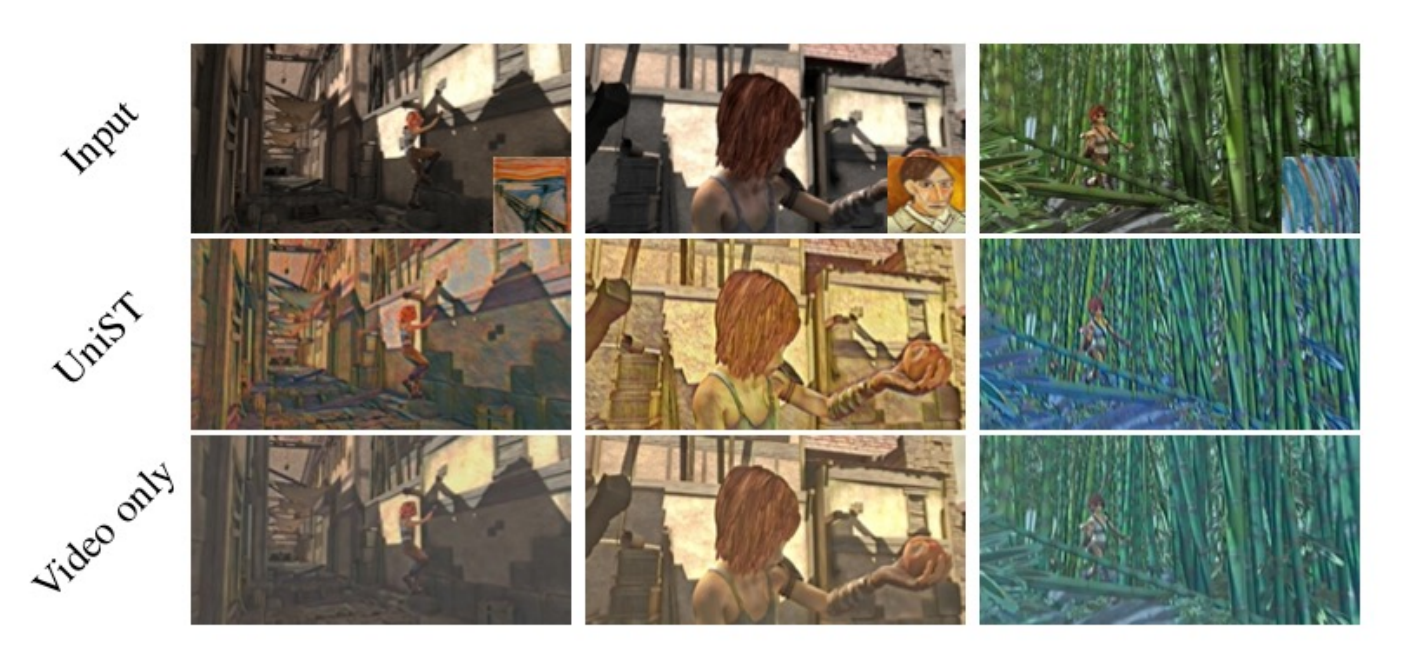}
\caption{Visualization of video style transfer with different training strategies. }
\label{fig:joint_video}
\end{figure}

\textbf{Video-image interaction module.}
Since images and videos have different domain features, we design a cross-attention domain-interaction module for joint learning of two domains for better content knowledge, which improves the overall stylization results. We conduct an experiment by removing the cross-attention interaction module. As in the 2nd, 3rd columns of Table \ref{tab:joint_learning}, both content preservation and style transfer are highly degraded without this module.

\begin{figure}[htb]
\centering
\includegraphics[width=0.48\textwidth]{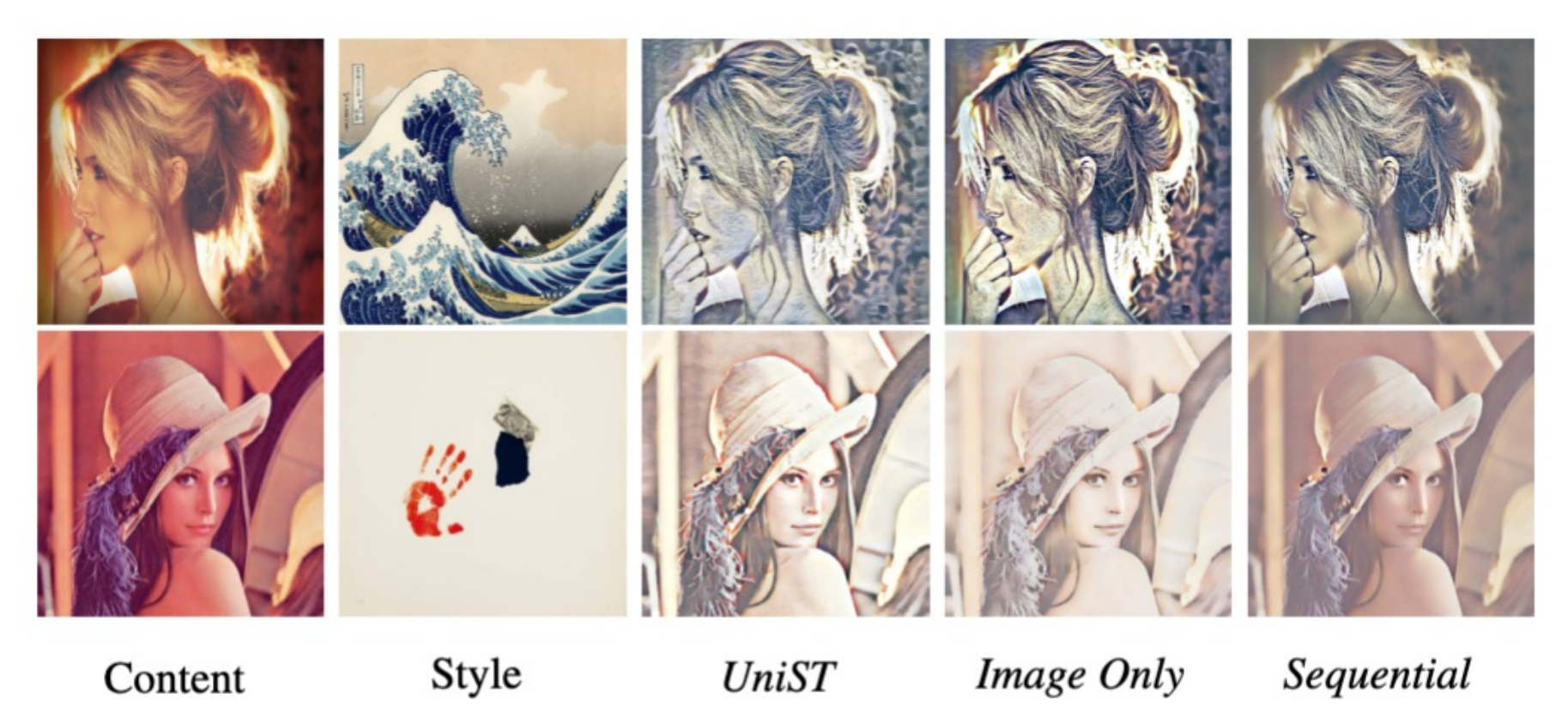}
\caption{Visualization of image style transfer with different training strategies.}
\label{fig:joint_image}
\end{figure}

To further demonstrate the superiority of UniST, we compare 3 training strategies without video-image interaction module. As shown in the 4th, 5th, 6th columns of Table \ref{tab:joint_learning}, UniST achieves the best performance compared with training separately and sequentially. Meanwhile, as shown in Figure \ref{fig:joint_video}, video style transfer is improved by learning complex appearances and textures from the image domain with UniST framework. Similarly, Figure \ref{fig:joint_image} shows UniST enhances image stylization effects with more vivid results.
 
\begin{table}[htb]
\renewcommand\arraystretch{1.2}
\scriptsize
\resizebox{0.47\textwidth}{!}{
\begin{tabular}{l|c c c c c}
     \Xhline{1pt}
     &UniST &Image+Video  & Video Only & Image Only & Sequential Training \\
     \Xhline{0.75pt}
        $ \mathcal{D}_C $ ($\downarrow$) &\bf{12.36} &15.44  &14.42    &19.29   &16.36  \\
     \Xhline{0.5pt}
       $ \mathcal{D}_S $ ($\downarrow$)  &\bf{0.46} & 0.69   &0.89   & 0.64   &0.69 \\
     \Xhline{0.75pt}
  \end{tabular}
  }
\caption{Effectiveness of joint learning framework. ``video only'' and ``image only'' indicate unimodal input for training, and ``Sequential learning'' means that we first train the network with one modality and then fine-tune it with another. Specifically, except for the UniST, the others all use the version without video-image interaction module.} 
\label{tab:joint_learning}
% \vspace{-1em}
\end{table}

%~~~~~~~~~~~~~~~~~~~~~~~~~~~~~

\renewcommand\arraystretch{1}
\begin{table}[!ht]\scriptsize
  \centering
    \begin{tabular}{c|ccc}
    \Xhline{1pt}
    &Image & Video &Image+Video \\
    \hline
    $ \mathcal{D}_C $($\downarrow$)       &   12.36   &n/a  &   12.36      \\
    $ \mathcal{D}_S $($\downarrow$)       &   0.46    & n/a &   0.46      \\
    Mean optical flow error($\downarrow$) &   n/a  &  3.86 &    3.86      \\
    Mean LPIPS($\downarrow$)              &   n/a    & 1.79  &   1.79      \\
    \Xhline{1pt}
    \end{tabular}
    \caption{ Metrics comparison of different input.}
  \label{tab:modality input}
\end{table}

\textbf{Inconsistent datasets concerns.}
In our method, we adopt two datasets from different modalities for learning. To ensure fairness, we have retrained five image style transfer models, while keeping the training dataset consistent with us. We provide the quantitative comparison in Table \ref{tab:consistent datasets}. Meanwhile, we conduct extra experiments by training the image-only UniST with different dataset scales in Table \ref{tab:datasets scales}. The results show simply using more images for image-only UniST brings no obvious improvements, while gains come from different modalities and beneficial interactions, and further illustrating the superiority of our model. 

 \begin{table}[!h]
  \centering
  \resizebox{0.47\textwidth}{!}{
  \begin{tabular}{l|c c c c c c}
    \Xhline{1pt}
    Methods & UniST &CCPL &StyleFormer &IEST &Stytr$^2$ &AdaAttN \\
    \Xhline{0.5pt}
     $ \mathcal{D}_C $($\downarrow$) & \bf{12.36} &14.32 ($\downarrow$ 0.34) &13.12 ($\downarrow$ 3.70)  &15.72 ($\uparrow$ 0.34) &12.87 ($\downarrow$ 0.80)  &14.76 ($\uparrow$ 0.22)\\
     $ \mathcal{D}_S $($\downarrow$) & \bf{0.46}  &0.85 ($\uparrow$ 0.14) &1.13 ($\uparrow$ 0.33)  &0.97 ($\downarrow$ 0.41) &0.61 ($\uparrow$ 0.11) &1.24 ($\uparrow$ 0.17)\\
    \Xhline{1pt}
  \end{tabular}
  }
  \caption{Quantitative comparison under consistent training dataset conditions. }
  \vspace{-1em}
  \label{tab:consistent datasets}
\end{table}

\renewcommand\arraystretch{1}
\begin{table}[!ht]
  \centering
    \resizebox{0.47\textwidth}{!}{
  \vspace{-4mm}
    \begin{tabular}{c|cccc}
    Training with different images     & \multicolumn{1}{l}{$\mathcal{D}_c$ $\downarrow$}   & \multicolumn{1}{c}{$\mathcal{D}_s$ $\downarrow$}  
    \\
          \hline
    UniST with videos and 60K images  & 12.36   &0.46    \\
    UniST using videos and 60K images without interaction &  15.44   & 0.69    \\
    Image-only UniST with 60K images  &19.80   &0.73    \\
    Image-only UniST with 70K images &19.80   &0.76    \\
    Image-only UniST with 80K images  &19.98   &0.71    \\

    \end{tabular}%
    }
 \caption{Comparison of UniST with image-only UniST using more training images.}
 \label{tab:datasets scales} %
  \vspace{-4mm}
\end{table}%

\begin{figure}[!ht]
\centering
\includegraphics[width=0.48\textwidth]{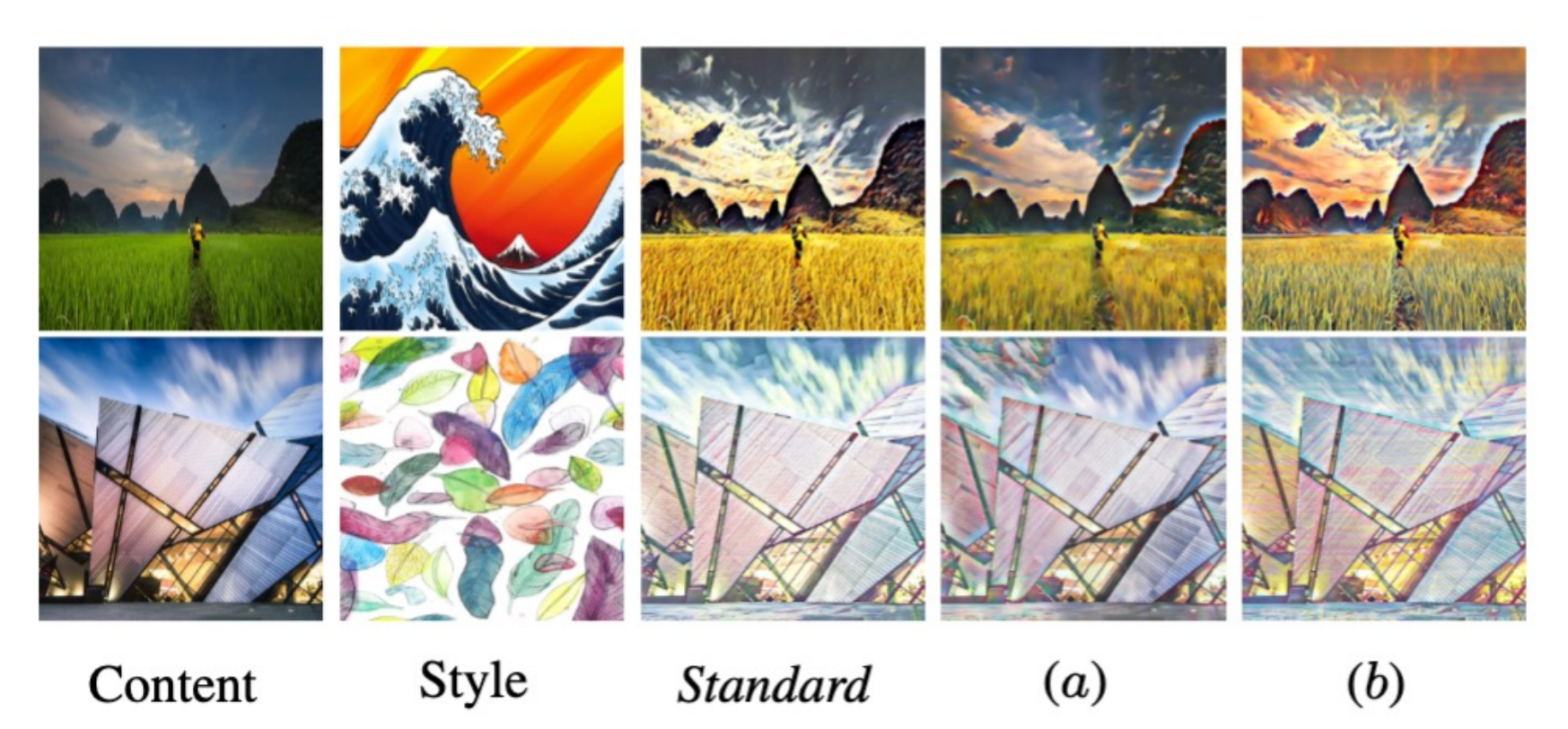}
\vspace{-10pt}
\caption{Ablation study of the AMSA layer. ``Standard'' is the normal version in Figure \ref{fig:axial}. ($a$) We uniformly take the output of the first MSA as the $Q$, $K$, $V$ of the second MSA. ($b$) We use the output of the first MSA as the $Q$ for the second MSA, while keeping $K$ and $V$ same as the first.}
\vspace{-1em}
\label{fig:attention}
\end{figure}

\textbf{Axial Multi-head Self-Attention.}
To verify the effectiveness of AMSA layer, we conduct two different design comparisons in Figure~\ref{fig:attention}. First, the information contained in the previous $K$, $V$ is crucial for the second MSA layer to learn contextual dependencies either within or across domains. 
Therefore, we uniformly take the output of the first MSA as the $Q$, $K$, $V$ for the second MSA. 
The results in 4th column show that it is difficult for the model to capture the relevance between content and style domains in this way. Second, the first axial information is necessary for the $V$ of the second MSA to maintain the stable stylization results. To demonstrate this, we use the output of the first MSA as the $Q$ for the second MSA, while keeping the $K$ and $V$ unchanged. As in the 5th column, the style patterns transferred are mixed with bar artifacts. From the results in the 3th column, the original design in Figure \ref{fig:axial} is necessary for our model to prevent side effects caused by axial attention.

%~~~~~~~~~~~~~~~~~~~~~~~~~~~~~~~~~~~~~~~~~
\vspace{-10pt}

%~~~~~~~~~~~~~~~~~~~~~~~~~~~~~~~~~~~~~~~~~~~~
\section{Conclusion}
In this work, we propose an unified style transfer framework, dubbed UniST, for image and video. The key is our novel domain interaction transformer that enables effective mutual feature learning from different modalities for enhancements. Besides, an axial multi-head attention is proposed to capture attentions either within or across the field efficiently. Experiment shows the mixing content input effectively improves stylization results via UniST.

\vspace{0.3em}
\noindent
{\bf Acknowledgement.} Libo Zhang was supported by Youth Innovation Promotion Association, CAS (2020111). Heng Fan and his employer received no financial support for research, authorship, and/or publication of this article.

{\small
\bibliographystyle{ieee_fullname}
\bibliography{egbib}

\begin{thebibliography}{10}\itemsep=-1pt

\bibitem{an2021artflow}
Jie An, Siyu Huang, Yibing Song, Dejing Dou, Wei Liu, and Jiebo Luo.
\newblock Artflow: Unbiased image style transfer via reversible neural flows.
\newblock In {\em {CVPR}}, pages 862--871, 2021.

\bibitem{butler2012naturalistic}
Daniel~J Butler, Jonas Wulff, Garrett~B Stanley, and Michael~J Black.
\newblock A naturalistic open source movie for optical flow evaluation.
\newblock In {\em {ECCV}}, pages 611--625, 2012.

\bibitem{DBLP:conf/iccv/ChenLYYH17}
Dongdong Chen, Jing Liao, Lu Yuan, Nenghai Yu, and Gang Hua.
\newblock Coherent online video style transfer.
\newblock In {\em {ICCV}}, pages 1114--1123, 2017.

\bibitem{DBLP:conf/nips/ChenZWZZLXL21}
Haibo Chen, Lei Zhao, Zhizhong Wang, Huiming Zhang, Zhiwen Zuo, Ailin Li, Wei
  Xing, and Dongming Lu.
\newblock Artistic style transfer with internal-external learning and
  contrastive learning.
\newblock In {\em NeurIPS}, pages 26561--26573, 2021.

\bibitem{cheng2023user}
Jiaxin Cheng, Yue Wu, Ayush Jaiswal, Xu Zhang, Pradeep Natarajan, and Prem
  Natarajan.
\newblock User-controllable arbitrary style transfer via entropy
  regularization.
\newblock 2023.

\bibitem{deng2009imagenet}
Jia Deng, Wei Dong, Richard Socher, Li-Jia Li, Kai Li, and Li Fei-Fei.
\newblock Imagenet: A large-scale hierarchical image database.
\newblock In {\em {CVPR}}, pages 248--255, 2009.

\bibitem{deng2021arbitrary}
Yingying Deng, Fan Tang, Weiming Dong, Haibin Huang, Chongyang Ma, and
  Changsheng Xu.
\newblock Arbitrary video style transfer via multi-channel correlation.
\newblock In {\em AAAI}, pages 1210--1217, 2021.

\bibitem{deng2020arbitrary}
Yingying Deng, Fan Tang, Weiming Dong, Wen Sun, Feiyue Huang, and Changsheng
  Xu.
\newblock Arbitrary style transfer via multi-adaptation network.
\newblock In {\em {ACM MM}}, pages 2719--2727, 2020.

\bibitem{deng2021stytr}
Yingying Deng, Fan Tang, Xingjia Pan, Weiming Dong, Chongyang Ma, and
  Changsheng Xu.
\newblock Stytr{\^{}}2: Unbiased image style transfer with transformers.
\newblock {\em CoRR}, abs/2105.14576, 2021.

\bibitem{DBLP:journals/spm/DokmanicPRV15}
Ivan Dokmanic, Reza Parhizkar, Juri Ranieri, and Martin Vetterli.
\newblock Euclidean distance matrices: Essential theory, algorithms, and
  applications.
\newblock {\em {IEEE} Signal Process. Mag.}, 32(6):12--30, 2015.

\bibitem{DBLP:conf/iclr/DosovitskiyB0WZ21}
Alexey Dosovitskiy, Lucas Beyer, Alexander Kolesnikov, Dirk Weissenborn,
  Xiaohua Zhai, Thomas Unterthiner, Mostafa Dehghani, Matthias Minderer, Georg
  Heigold, Sylvain Gelly, Jakob Uszkoreit, and Neil Houlsby.
\newblock An image is worth 16x16 words: Transformers for image recognition at
  scale.
\newblock In {\em {ICLR}}. OpenReview.net, 2021.

\bibitem{DBLP:conf/wacv/GaoLY020}
Wei Gao, Yijun Li, Yihang Yin, and Ming{-}Hsuan Yang.
\newblock Fast video multi-style transfer.
\newblock In {\em {WACV}}, pages 3211--3219, 2020.

\bibitem{DBLP:conf/cvpr/GatysEB16}
Leon~A. Gatys, Alexander~S. Ecker, and Matthias Bethge.
\newblock Image style transfer using convolutional neural networks.
\newblock In {\em {CVPR}}, pages 2414--2423, 2016.

\bibitem{DBLP:journals/corr/abs-2104-05704}
Ali Hassani, Steven Walton, Nikhil Shah, Abulikemu Abuduweili, Jiachen Li, and
  Humphrey Shi.
\newblock Escaping the big data paradigm with compact transformers.
\newblock {\em CoRR}, abs/2104.05704, 2021.

\bibitem{DBLP:conf/cvpr/HuangWLMJZLL17}
Haozhi Huang, Hao Wang, Wenhan Luo, Lin Ma, Wenhao Jiang, Xiaolong Zhu, Zhifeng
  Li, and Wei Liu.
\newblock Real-time neural style transfer for videos.
\newblock In {\em {CVPR}}, pages 7044--7052, 2017.

\bibitem{huang2017arbitrary}
Xun Huang and Serge Belongie.
\newblock Arbitrary style transfer in real-time with adaptive instance
  normalization.
\newblock In {\em {ICCV}}, pages 1501--1510, 2017.

\bibitem{DBLP:conf/iccv/IgnatovKTVG17}
Andrey Ignatov, Nikolay Kobyshev, Radu Timofte, Kenneth Vanhoey, and Luc~Van
  Gool.
\newblock Dslr-quality photos on mobile devices with deep convolutional
  networks.
\newblock In {\em {IEEE} International Conference on Computer Vision, {ICCV}
  2017, Venice, Italy, October 22-29, 2017}, pages 3297--3305. {IEEE} Computer
  Society, 2017.

\bibitem{johnson2016perceptual}
Justin Johnson, Alexandre Alahi, and Li Fei-Fei.
\newblock Perceptual losses for real-time style transfer and super-resolution.
\newblock In {\em {ECCV}}, pages 694--711. Springer, 2016.

\bibitem{karras2019style}
Tero Karras, Samuli Laine, and Timo Aila.
\newblock A style-based generator architecture for generative adversarial
  networks.
\newblock In {\em {CVPR}}, pages 4401--4410, 2019.

\bibitem{kong2023exploring}
Xiaoyu Kong, Yingying Deng, Fan Tang, Weiming Dong, Chongyang Ma, Yongyong
  Chen, Zhenyu He, and Changsheng Xu.
\newblock Exploring the temporal consistency of arbitrary style transfer: A
  channelwise perspective.
\newblock {\em IEEE Transactions on Neural Networks and Learning Systems},
  2023.

\bibitem{li2019learning}
Xueting Li, Sifei Liu, Jan Kautz, and Ming-Hsuan Yang.
\newblock Learning linear transformations for fast image and video style
  transfer.
\newblock In {\em {CVPR}}, pages 3809--3817, 2019.

\bibitem{li2017universal}
Yijun Li, Chen Fang, Jimei Yang, Zhaowen Wang, Xin Lu, and Ming-Hsuan Yang.
\newblock Universal style transfer via feature transforms.
\newblock {\em {NeurIPS}}, 30, 2017.

\bibitem{lin2021drafting}
Tianwei Lin, Zhuoqi Ma, Fu Li, Dongliang He, Xin Li, Errui Ding, Nannan Wang,
  Jie Li, and Xinbo Gao.
\newblock Drafting and revision: Laplacian pyramid network for fast
  high-quality artistic style transfer.
\newblock In {\em {CVPR}}, pages 5141--5150, 2021.

\bibitem{lin2014microsoft}
Tsung-Yi Lin, Michael Maire, Serge Belongie, James Hays, Pietro Perona, Deva
  Ramanan, Piotr Doll{\'a}r, and C~Lawrence Zitnick.
\newblock Microsoft coco: Common objects in context.
\newblock In {\em {ECCV}}, pages 740--755, 2014.

\bibitem{liu2021adaattn}
Songhua Liu, Tianwei Lin, Dongliang He, Fu Li, Meiling Wang, Xin Li, Zhengxing
  Sun, Qian Li, and Errui Ding.
\newblock Adaattn: Revisit attention mechanism in arbitrary neural style
  transfer.
\newblock In {\em {ICCV}}, pages 6649--6658, 2021.

\bibitem{luo2022consistent}
Xuan Luo, Zhen Han, Lingkang Yang, and Lingling Zhang.
\newblock Consistent style transfer.
\newblock {\em arXiv preprint arXiv:2201.02233}, 2022.

\bibitem{park2019arbitrary}
Dae~Young Park and Kwang~Hee Lee.
\newblock Arbitrary style transfer with style-attentional networks.
\newblock In {\em {CVPR}}, pages 5880--5888, 2019.

\bibitem{phillips2011wiki}
Fred Phillips and Brandy Mackintosh.
\newblock Wiki art gallery, inc.: A case for critical thinking.
\newblock {\em Issues in Accounting Education}, 26(3):593--608, 2011.

\bibitem{sheng2018avatar}
Lu Sheng, Ziyi Lin, Jing Shao, and Xiaogang Wang.
\newblock Avatar-net: Multi-scale zero-shot style transfer by feature
  decoration.
\newblock In {\em {CVPR}}, pages 8242--8250, 2018.

\bibitem{simonyan2014very}
Karen Simonyan and Andrew Zisserman.
\newblock Very deep convolutional networks for large-scale image recognition.
\newblock {\em arXiv preprint arXiv:1409.1556}, 2014.

\bibitem{vaswani2017attention}
Ashish Vaswani, Noam Shazeer, Niki Parmar, Jakob Uszkoreit, Llion Jones,
  Aidan~N Gomez, {\L}ukasz Kaiser, and Illia Polosukhin.
\newblock Attention is all you need.
\newblock {\em {NIPS}}, 30, 2017.

\bibitem{wang2020axial}
Huiyu Wang, Yukun Zhu, Bradley Green, Hartwig Adam, Alan Yuille, and
  Liang-Chieh Chen.
\newblock Axial-deeplab: Stand-alone axial-attention for panoptic segmentation.
\newblock In {\em {ECCV}}, pages 108--126, 2020.

\bibitem{DBLP:journals/corr/abs-2211-15313}
Zhizhong Wang, Lei Zhao, Zhiwen Zuo, Ailin Li, Haibo Chen, Wei Xing, and
  Dongming Lu.
\newblock Microast: Towards super-fast ultra-resolution arbitrary style
  transfer.
\newblock {\em CoRR}, abs/2211.15313, 2022.

\bibitem{Ranger}
Less Wright.
\newblock Ranger - a synergistic optimizer.
\newblock \url{https://github.com/lessw2020/Ranger-Deep-Learning-Optimizer},
  2019.

\bibitem{DBLP:conf/iccv/WuXCLDY021}
Haiping Wu, Bin Xiao, Noel Codella, Mengchen Liu, Xiyang Dai, Lu Yuan, and Lei
  Zhang.
\newblock Cvt: Introducing convolutions to vision transformers.
\newblock In {\em {ICCV}}, pages 22--31. {IEEE}, 2021.

\bibitem{wu2020preserving}
Xinxiao Wu and Jialu Chen.
\newblock Preserving global and local temporal consistency for arbitrary video
  style transfer.
\newblock In {\em {ACM MM}}, pages 1791--1799, 2020.

\bibitem{DBLP:conf/iccv/WuHS021}
Xiaolei Wu, Zhihao Hu, Lu Sheng, and Dong Xu.
\newblock Styleformer: Real-time arbitrary style transfer via parametric style
  composition.
\newblock In {\em {ICCV}}, pages 14598--14607. {IEEE}, 2021.

\bibitem{DBLP:journals/corr/abs-2207-04808}
Zijie Wu, Zhen Zhu, Junping Du, and Xiang Bai.
\newblock {CCPL:} contrastive coherence preserving loss for versatile style
  transfer.
\newblock {\em CoRR}, abs/2207.04808, 2022.

\bibitem{DBLP:conf/wacv/XiaXLSCKC21}
Xide Xia, Tianfan Xue, Wei{-}Sheng Lai, Zheng Sun, Abby Chang, Brian Kulis, and
  Jiawen Chen.
\newblock Real-time localized photorealistic video style transfer.
\newblock In {\em {WACV}}, pages 1088--1097, 2021.

\bibitem{DBLP:journals/corr/abs-2103-11816}
Kun Yuan, Shaopeng Guo, Ziwei Liu, Aojun Zhou, Fengwei Yu, and Wei Wu.
\newblock Incorporating convolution designs into visual transformers.
\newblock {\em CoRR}, abs/2103.11816, 2021.

\bibitem{DBLP:conf/cvpr/ZhangIESW18}
Richard Zhang, Phillip Isola, Alexei~A. Efros, Eli Shechtman, and Oliver Wang.
\newblock The unreasonable effectiveness of deep features as a perceptual
  metric.
\newblock In {\em {CVPR}}, pages 586--595. Computer Vision Foundation / {IEEE}
  Computer Society, 2018.

\end{thebibliography}
}

\end{document}